\documentclass{article}

\usepackage[dvipdfmx]{graphicx}
\usepackage{PRIMEarxiv}

\usepackage[utf8]{inputenc} 
\usepackage[T1]{fontenc}    
\usepackage{hyperref}       
\usepackage{url}            
\usepackage{booktabs}       
\usepackage{amsfonts}       
\usepackage{nicefrac}       
\usepackage{microtype}      
\usepackage{lipsum}
\usepackage{fancyhdr}       
\usepackage{graphicx}       

\usepackage{amsmath}
\usepackage{diagbox}

\DeclareMathOperator*{\argmax}{arg\,max}

\def\baa{{ \boldsymbol a}}
\def\bb{{ \boldsymbol b}}

\def\bv{{ \boldsymbol v}}

\title{Estimation of User's World Model Using Graph2vec
\thanks{
This paper is an extended (English translated) version of ``T.Sakai, T.Horii, T.Nagai, Representation Learning of World Models and Estimation of World Model of Others Using Graph2vec, Journal of RSJ, 40, 2, pp.166-169, 2022,'' (in Japanese) ISSN 1884-7145, Print ISSN 0289-1824, https://doi.org/10.7210/jrsj.40.166
}}

\author{
  Tatsuya Sakai, ~~~~~Takayuki Nagai \\
  Graduate School of Engineering Science, Osaka University \\
  1-3, Machikaneyama, Toyonaka, Osaka, Japan\\
  \texttt{nagai@sys.es.osaka-u.ac.jp} \\
}

\begin{document}
\maketitle

\begin{abstract}
To obtain advanced interaction between autonomous robots and users, robots should be able to distinguish their state space representations (i.e., world models).
Herein, a novel method was proposed for estimating the user's world model based on queries.
In this method, the agent learns the distributed representation of world models using graph2vec and generates concept activation vectors that represent the meaning of queries in the latent space. 
Experimental results revealed that the proposed method can estimate the user's world model more efficiently than the simple method of using the ``AND'' search of queries. 

\end{abstract}


\section{Introduction}
Autonomous robots are increasingly being used in numerous applications.
Currently, they assist humans in performing tasks by executing commands.
For autonomous robots performing sophisticated decisions, the blind execution of commands is not always the best strategy.
Moreover, in many situations, fully executing commands is difficult.
These autonomous robots should be able to explain the reasons for their decisions to gain user trust.
Explainable autonomous robots (XAR) are defined as robots that have these explanatory capabilities. The following four requirements have been identified for the XARs \cite{Sakai_survey}:

\begin{itemize}
\item[(1)] {Owning an interpretable decision-making space}
\item[(2)] {Estimation of the model of others}
\item[(3)] {Estimation of the information required for a user to estimate the policy of the robot} 
\item[(4)] {Presentation to the user of explanation} 
\end{itemize}

\begin{figure}[t]
\centering
    \includegraphics[width=0.75\linewidth]{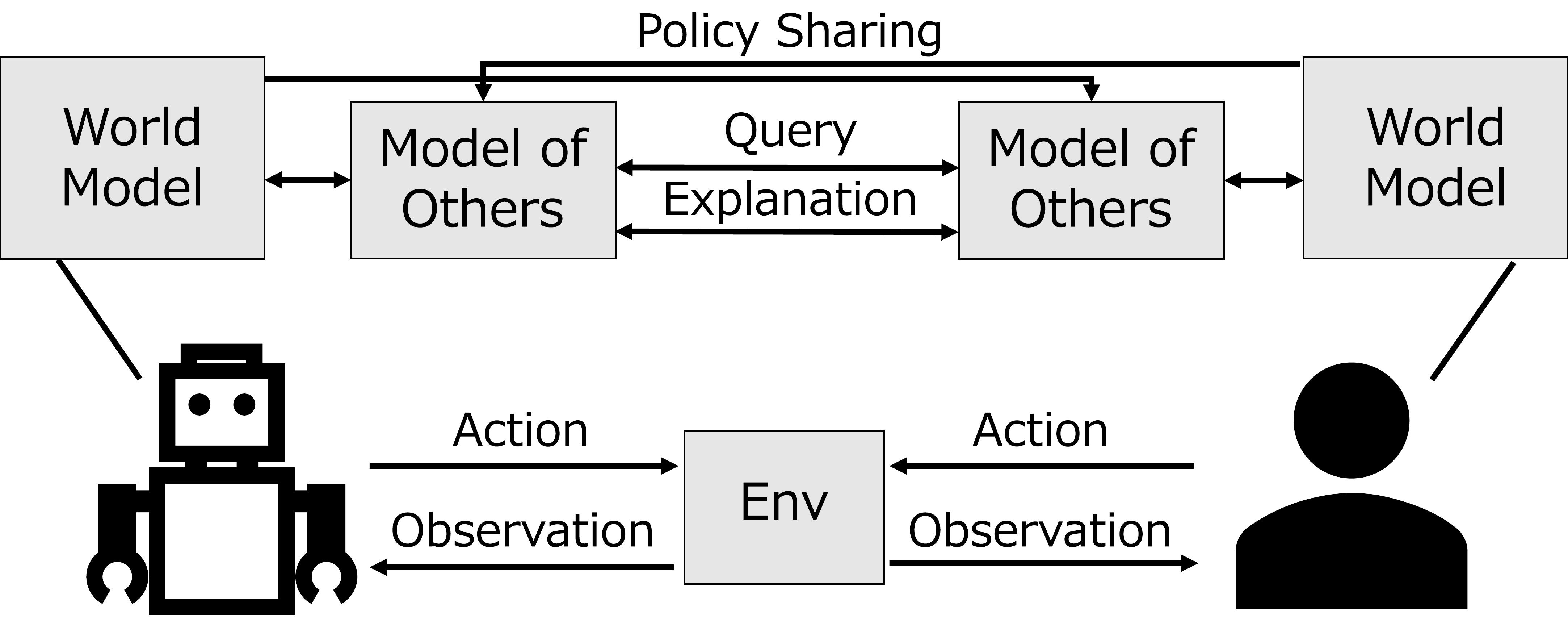}
    \caption{Explanation process as communication. To clarify elements of the explanation, observations/actions, and interactions between world models of others are segregated in the figure.} 
    \label{fig:intro1}
\end{figure}

This explanation mechanism is a mutual process between the robot and the user and is displayed in Fig. \ref{fig:intro1}.
The world model refers to the correspondence between actions and state changes, that is, the internal model \cite{Ha} that represents the dynamics of the environment, and we do not distinguish between the world model and the environment in this paper. 
``Policy sharing'' in Fig.\ref{fig:intro1} is a spontaneous presentation of information of a policy (e.g., presentation of a sequence of actions to be taken), and an explanation is generated when a query to this information presentation is requested by others.

Among the four requirements, the estimation of the other's world model is particularly crucial for providing a user-specific explanation. 
In the context of human--robot interaction, the importance of estimating the user's internal state has already been recognized. 
Gao {\it et al.} \cite{Gao} and Clair {\it et al.} \cite{Clair} proposed a framework for estimating plausible action strategies based on the actions and interaction history of users. 
Huang {\it et al.} \cite{Huang} and Lage {\it et al.} \cite{Lage} focused on restoring explanations to policies and advocated the importance of appropriately estimating the restoration algorithm of users requesting explanation.
These studies have focused on internal states, particularly policies, and planning algorithms. 
However, real-world robots are designed to exhibit desired behavior in terms of algorithms and policies, and they share the same final objective with their users. 
In these situations, the results of action decisions typically stem from discrepancies in environmental awareness.

\begin{figure}[t]
\centering
    \includegraphics[width=0.8\linewidth]{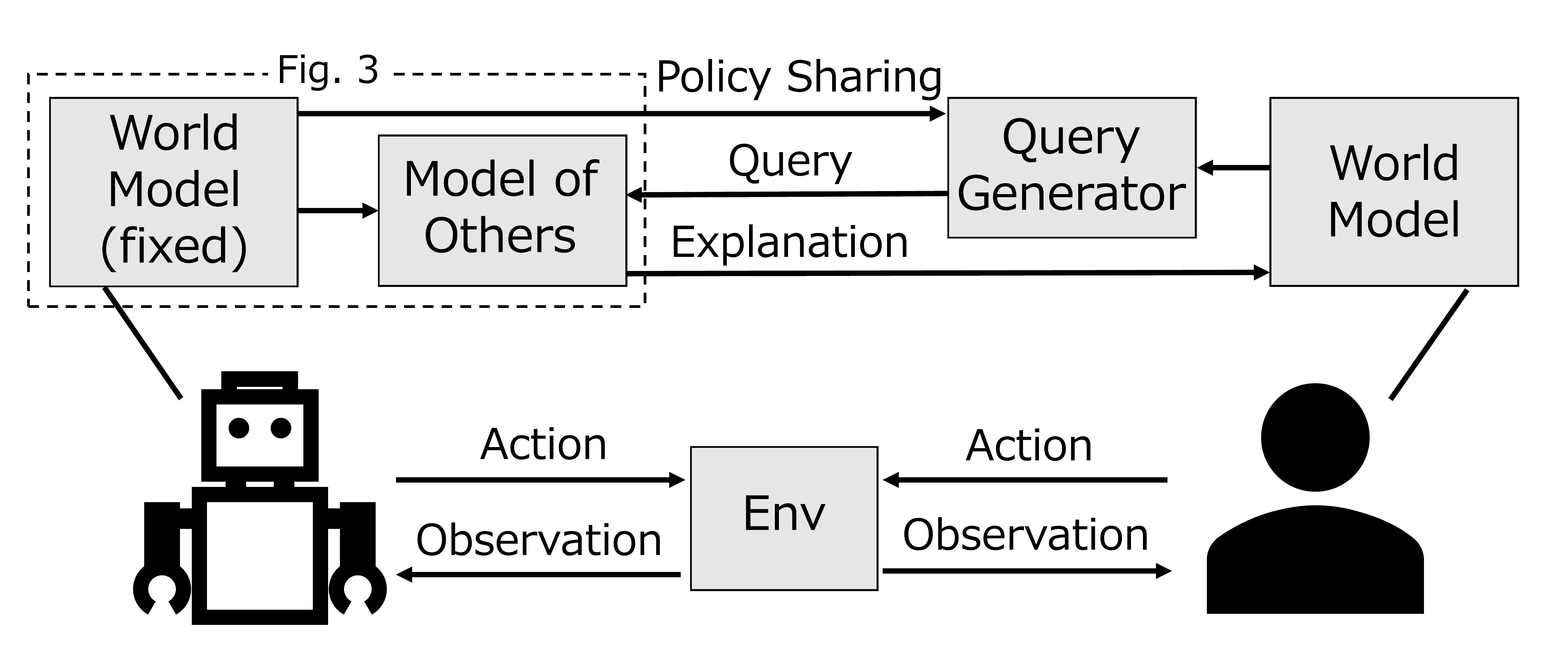}
    \caption{Simplified explanation process considered in this paper. We only consider unidirectional explanation process from robot to user.} 
    \label{fig:intro2}
\end{figure}

In this study, a novel method was proposed for estimating the user's world model based on the robot's world model and the questions (queries) posed by the user.
The proposed method can identify differences in the world models and generate an explanation that resolves the discrepancy between the perception of the environment by the robot and the user.
In this study, we simplified the mechanism of Fig.\ref{fig:intro1} as in Fig.\ref{fig:intro2} \footnote{When the model of others outputs an explanation, the content of the explanation is determined by considering information received from the world model.} and focused on estimating the world model of the explained person as the user's world model.


\section{Proposed method}
\begin{figure}[t]
\centering
    \includegraphics[width=0.8\linewidth]{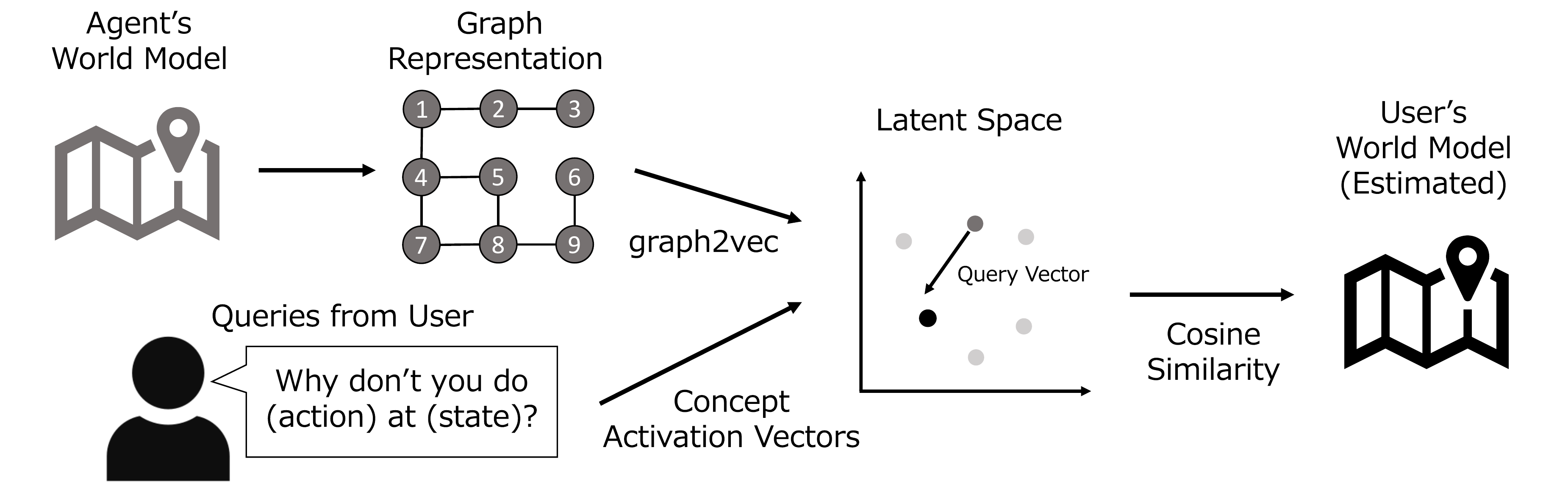}
    \caption{Schematic of our proposed method.}
    \label{fig:overall}
\end{figure}

In the proposed method, the world model of the user is estimated by the following procedure (Fig. \ref{fig:overall}).

\begin{itemize}
\item[(1)] {\bf Acquisition of a distributed representation of the world model:} 
A distributed representation of each world model is obtained using a graph-structured world model. 

\item[(2)] {\bf Acquisition of the query vector:}
Based on the query given by the user, the system acquires a direction vector that represents the meaning of the query in the representation space of the world model. 

\item[(3)] {\bf Estimation of user's world model:} 
The distributed representations of the world model and query vector are used to estimate the user's world model based on cosine similarity.

\end{itemize}

The robot and user are assumed to share the same state space and measures; only the connection relation between states is assumed to be unshared.
As a distributed representation of the world model, the parameters of a model representing a continuous state space, for example, \cite{Ha}, could be used.
However, presenting the differences of the world model to the user requires the discretization of the state transition structure; the parameter space of the model does not necessarily represent the similarity of the state transition structure of the world model. Therefore, the world model is considered to be a discretized graph for obtaining a distributed representation.
The generation of explanations is outside the scope of this study.

\begin{figure}[t]
\centering
\includegraphics[width=0.8\linewidth]{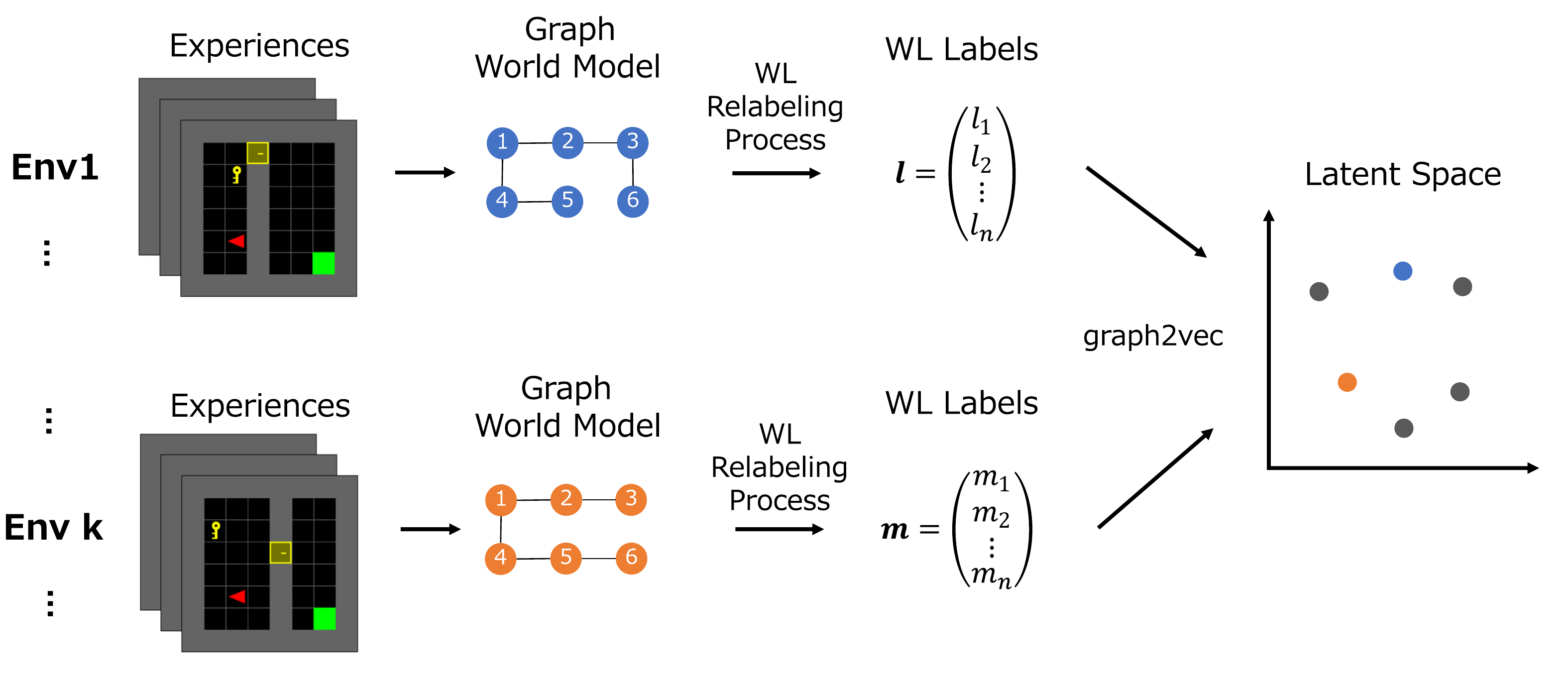}
\caption{Learning of a distributed representation of the world model. The experience on the environment is used to obtain graph-based world models. Then, after converting to WL labels, a distributed representation of each environment is obtained.}
\label{fig:graph2vec}
\end{figure}

\subsection{Acquisition of a distributed representation of the world model}
The learning process of a distributed representation of the world model is shown in Fig. \ref{fig:graph2vec}. 
The robot learns a representation space representing the similarity of environments based on its experiences.
First, the robot acquires an undirected graph representing the state transitions of each environment; simultaneously, it acquires policies through reinforcement learning.
Specifically, the robot assumes that the states whose transitions were observed during the policy learning search are adjacent to each other and adds edges\footnote{This method is assumed to be used in a discrete state space; discretization of the state space is required for application to a real robot. However, this measure is beyond the scope of this study. For discretization, the method proposed in \cite{Zhang}, wherein the policy implications of each state is considered, can be used.
}.

After acquiring the undirected graphs of all environments, graph2vec is applied to acquire the distributed representation of the graph of each environment \cite{Narayanan}.
Graph2vec is a method in which doc2vec \cite{Le} is used to acquire distributed representations of graphs, and instead of predicting the occurrence of words, the occurrence of labels that represent each subgraph is presented.
Labels representing subgraphs are obtained using the Weisfeiler--Lehman (WL) relabeling process \cite{Shervashidze}.
In this process, the label of the next layer is determined by considering the labels of neighboring nodes.
Higher layer labels represent more global information regarding the graph.

Graph2vec allows graphs in which the same subgraphs occur, that is, graphs in environments with similar state transition structures, to be embedded closer together in the representation space. 
Efficient search based on user queries becomes possible by acquiring the representation space of the world model in advance.

When acquiring a world model, we explicitly provide an environment label to indicate the environment wherein the experience occurs. 
If no environment label is given, a world model is obtained using temporally continuous experiences.
Furthermore, we consider that models with similar distributed representations represent the same environment and merging multiple models may be effective in obtaining a world model with higher accuracy.

\subsection{Acquisition of the query vector}
Based on the query given by the user, the robot acquires a direction vector that represents the meaning of the query in the representation space of the world model.
Kim {\it et al.} \cite{Kim} focused on the middle layer of the neural network and generated concept vectors (CAV: concept activation vectors) by calculating the difference in latent representations when features that satisfy a specific concept and features that do not satisfy the concept are input. 
In this study, this method was applied to define a query vector $\bv_{query}$ as Eq. (\ref{CAV}) by considering the difference of distributed representations between environments that satisfy the query and those that do not. 
The query assumes the form ``action $a_{query}$ should be selected in state $s_{query}$.''

\begin{equation}
\label{CAV}
\bv_{query}=\bv_{pos}-\bv_{neg}, 
\end{equation}

where

\begin{equation}
\bv_{pos}=\frac{\sum_{i}
\bv_{i}\cdot
P(a_{query}|\bv_{i},s_{query})}{\sum_{i} P(a_{query}|\bv_{i},s_{query})},~~~
\bv_{neg}=\frac{\sum_{i} \bv_{i}\cdot
(1-P(a_{query}|\bv_{i},s_{query}))}{\sum_{i} (1-P(a_{query}|\bv_{i},s_{query}))}.
\end{equation}

Here, $\bv_{i}$ represents a distributed representation of the $i$-th environment. 
To correspond to a policy in which actions are selected probabilistically, the probability value of selecting an action is used as the coefficient of each distributed representation $\bv_{i}$. 
If necessary, the coefficients can be expressed as the binary values of $\{0,1\}$.
Note that if the state $S_{query}$ is not included in the undirected graph of the environment $i$, the CAV is calculated by excluding $\bv_{i}$\footnote{When estimating the user's world model, the likelihood $S(\bv_{i},\bv_{obs}, V_q)$ is computed for all environments $i$, including those that do not contain the state $s_{query}$.}.

\subsection{Estimation of user's world model}
Using the distributed representation of the world model and the query vector, the user's world model was estimated using cosine similarity.
The likelihood of an environment $i$ as an estimated environment is expressed by Eq. (\ref{eval}). 

\begin{equation}
\label{eval}
S(\bv_{i},\bv_{obs}, V_q)= {\sum_{j} similarity(\bv_{query}^{j}, \bv_{i}-\bv_{obs})} -\lambda \cdot distance(\bv_{i}, \bv_{obs}),
\end{equation}

where $\bv_{obs}$ is a distributed representation of the environment currently observed by the agent, $V_q$ is any number of query vectors, and $\bv_{query}^{j} \in V_q$ is the $j$-th query vector. 
Furthermore, $similarity(\baa, \bb)$ and $distance(\baa,\bb)$ are functions that output the cosine similarity $[-1,1]$ and the distance between vectors $\baa$ and $\bb$ in the representation space, respectively. 

When reasoning in a real environment, the robot and the user observe almost the same environment. Therefore, because their world models are similar, the similarity between the direction of each environment and the direction of the query vector, as seen from $\bv_{obs}$, is used as the estimation criterion. 
The coefficient $\lambda$ is a hyperparameter that determines how much distance between vectors is considered and represents the strength of the assumption that the observed environment of the robot and the user are similar. 

Using the definitions, the world model of others to be estimated is expressed by Eq. (\ref{others}).

\begin{equation}
\label{others}
Env\_est=\argmax_i S(\bv_{i},\bv_{obs}, V_q)
\end{equation}

In this study, we assume that the importance of all queries are equivalent and designed the evaluation function $S$ as the sum of the similarities for each query vector $\bv_{query}^{j}$. 
However, in the real world, the importance of each query may differ, in which case the similarity should be multiplied by a coefficient $\rho^{j}$. 

\subsubsection{User vectors}
In addition to the query vector, the user vector that represents ``what kind of environment with the distributed representation the user is likely to retain as a world model'' can also be defined. 
The user vector is a vector that represents how the user tends to misunderstand the environment and plays a role in adding this tendency to the result of the user's world model estimation.

\begin{equation}
\label{user_vec}
\bv_{user}=\bv_{u\_pos}-\bv_{u\_neg}, 
\end{equation}

where 

\begin{equation}
\bv_{u\_pos}=\frac{\sum_{i}
\bv_{i}\cdot
P(\bv_{i})}{\sum_{i} P(\bv_{i})},~~~~
\bv_{u\_neg}=\frac{\sum_{i}
\bv_{i}\cdot
(1-P(\bv_{i}))}{\sum_{i} (1-P(\bv_{i}))}.
\end{equation}

$P(\bv_{i})$ is the probability that the user estimates the environment corresponding to the distributed representation $\bv_{i}$ as the current world model when no information about the current environment is given to the user. In this paper, $P(\bv_{i})$ is assumed to be known, and the estimation of $P(\bv_{i})$ is outside the scope of this research.

\subsection{Explanation by language}
Using a pre-trained language vector Eq. (\ref{word_vec}) in the representation space, the relationship between the world model maintained by the agent and the user's world model can also be explained by language.

\begin{equation}
\label{word_vec}
\bv_{word}(x)=\mbox{average}(\bv_{m}-\bv_{n}), 
\end{equation}

where $\bv_{m}$ and $\bv_{n}$ are distributed representations of environment pairs whose relation is represented by the language $x$. 
By averaging the differences of distributed representations of environment pairs $\bv_{m}$ and $\bv_{n}$ that satisfy a specific language description $x$, we can obtain a semantic vector represented by the language description $x$ in the representation space.

Using the language vector $\bv_{word}$, the language describing the relationship between the world model maintained by the agent and the user's world model is represented by Eq. (\ref{relation}).

\begin{equation}
\label{relation}
Explanation=\argmax_x similarity(\bv_{word}(x), \bv_{Env\_est} - \bv_{obs})).
\end{equation}

By means of Eq. (\ref{relation}), the language $x$ that represents the closest relationship between the current world models is selected as the explanation.

\section{Experiment}
\begin{figure}[t]
\centering
    \includegraphics[width=0.75\linewidth]{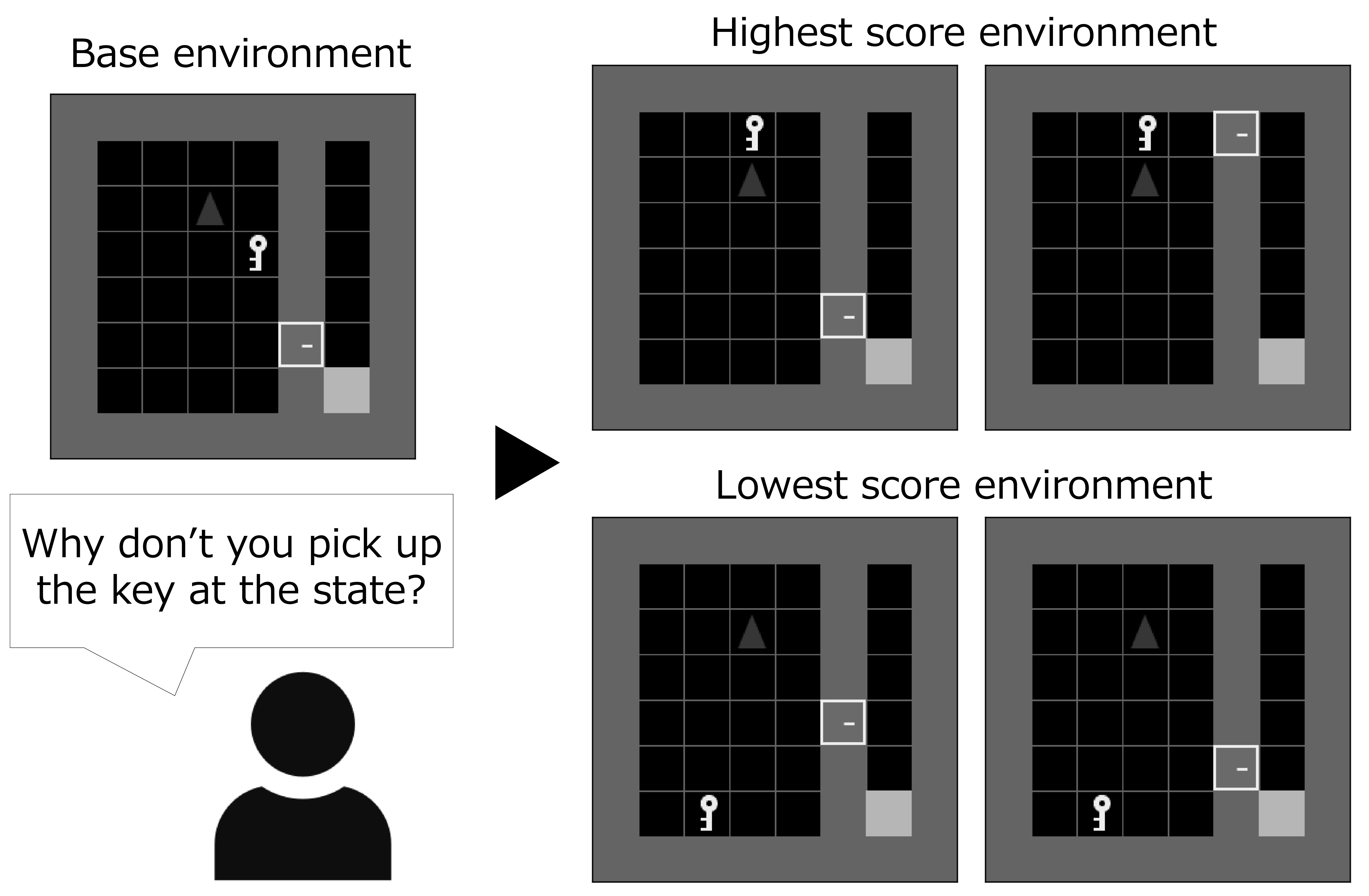}
    \caption{Estimation results of the user's world model.}
    \label{fig:exp1-2}
\end{figure}

The proposed method was applied to an agent that acquires action strategies by proximal policy optimization (PPO) in a simulation environment, and its usefulness was evaluated.
A partially modified version of the grid environment \cite{gym_minigrid} with multiple objects was used for the experiments (Fig. \ref{fig:exp1-2}). 
In this environment, the agent (triangle) obtains a reward by taking a key, opening a door, and reaching a goal in the lower right corner. 
The position of the goal remains unchanged, but the positions of the key and the door change every trial. 
The agent has five actions, namely go straight, turn left, turn right, take the key from the grid in front of it, and open the door. 
The agent observes the absolute position of the key ($x,y$ coordinates), the absolute position of the door ($x,y$ coordinates), its own absolute position ($x,y$ coordinates) and orientation, holding/not holding the key, and opening/closing the door, for a total of nine dimensions.

\subsection{Experiment 1: Acquisition of a distributed representation of the world model}
\begin{figure}[t]
\centering
    \includegraphics[width=0.55\linewidth]{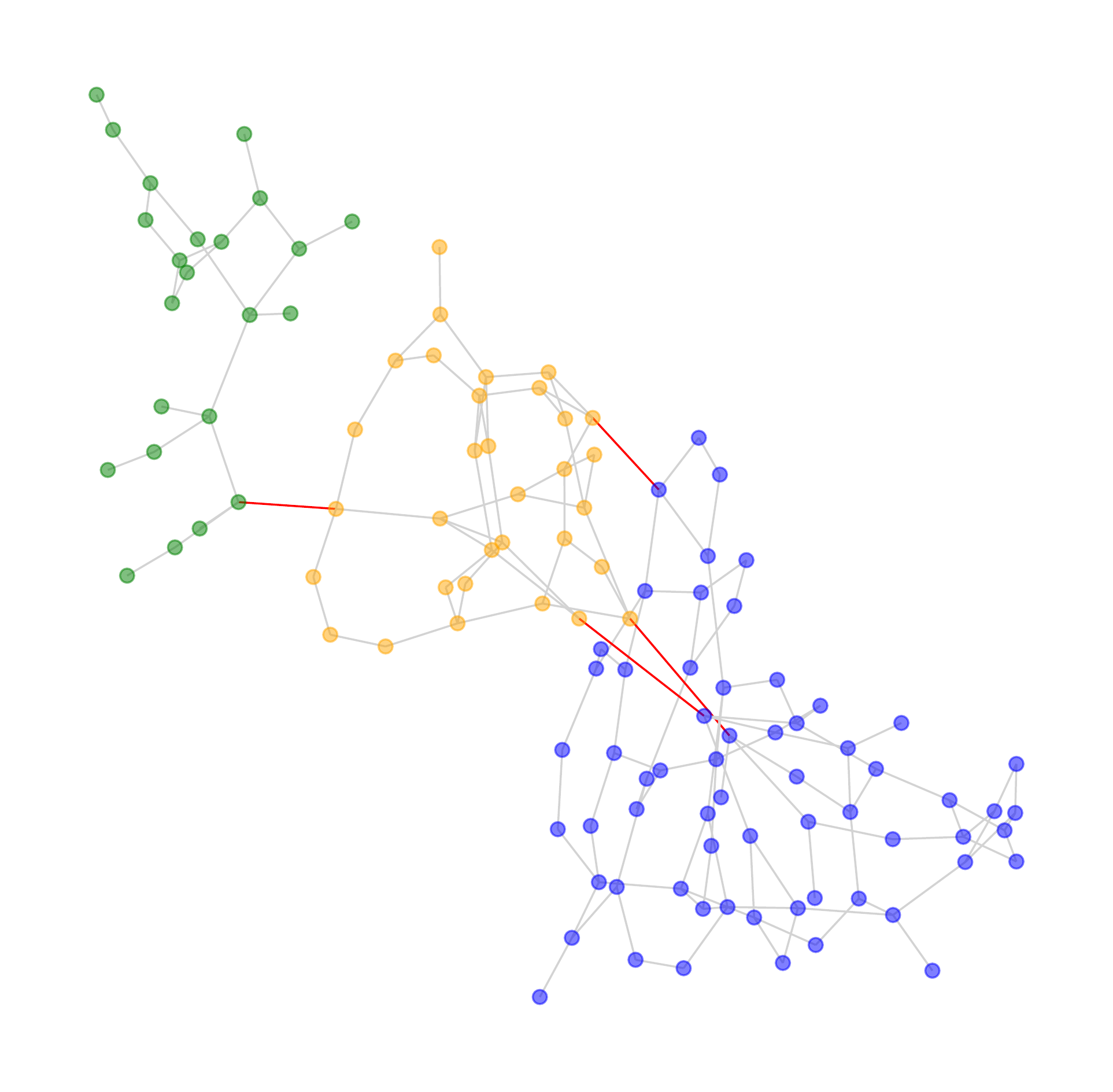}
    \caption{An example graph of the acquired world model. 
    The blue nodes represent the state before the key acquisition. The orange nodes represent the state after the key acquisition when the door is not open, and the green nodes represent the state when the door is open.} 
    \label{fig:worldmodel} 
\end{figure}

A graph representing the state transitions of each environment was obtained simultaneously with the learning of policies by PPO, and graph2vec was used to obtain a 16-dimensional distributed representation. 
An example of the acquired graph is shown in Fig.\ref{fig:worldmodel}.
There are three edges that transition from the group of states before key acquisition to the group of states after key acquisition because keys can be acquired from three directions.
On the other hand, there is only one edge that transitions from the group of states where the door is not open to the group of states where the door is open, because the door can be opened from only one state.

\begin{figure}[t]
\centering
    \includegraphics[width=0.7\linewidth]{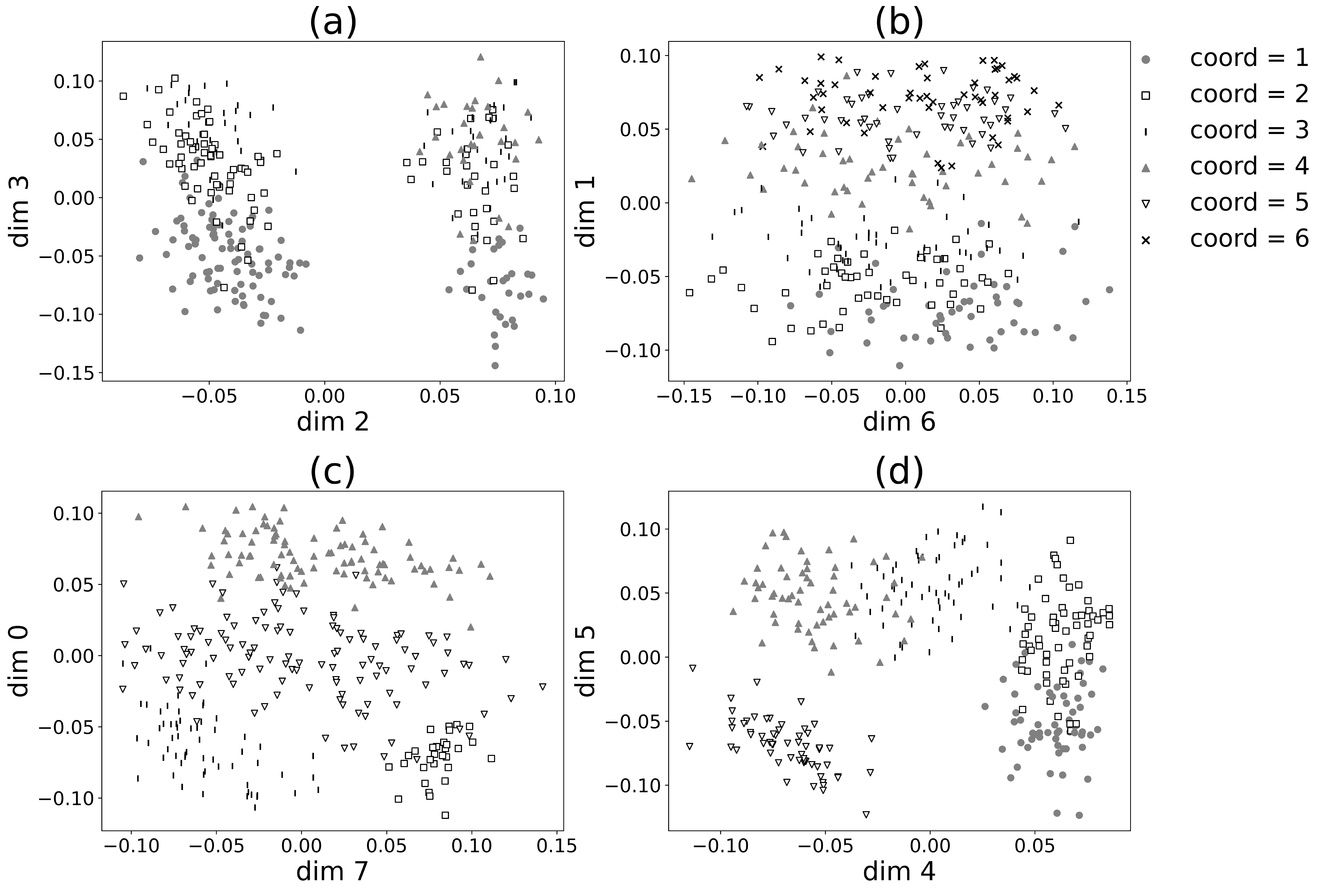}
    \caption{Results of latent space visualization. Each data point is illustrated according to (a) X-coordinate of the key, (b) Y-coordinate of the key, (c) X-coordinate of the door, and (d) Y-coordinate of the door.} 
    \label{fig:emb}
\end{figure}

The representation space was compressed to eight dimensions using independent component analysis, and the visualized results are displayed in Fig. \ref{fig:emb}.
In this experiment, the absolute positions of the key and door were used as environment labels, and the five-dimensional observations excluding them were used as node features.
The experimental results revealed that clusters are formed in the representation space based on the absolute positions of the key and door.
In particular, clusters related to the position of the door are apparent, which can be attributed to the fact that the surrounding state transition relationship changes considerably compared with that of the key.
In this experiment, the absolute positions of keys and doors are used only for identifying the environmental graph to be updated and are not embedded in the graph itself. Therefore, graph2vec created a representation space that appropriately reflects the differences in the location information of keys and doors expressed in the state transition structure.

\begin{table}[t]
\begin{center}
\caption{Cumulative frequency and average value of the order in which the optimal environment appears. The number of trials is 100 for each method.}
\scalebox{1.0}{
\begin{tabular}{c|c c  c|| c}
   \diagbox{Method}{Order}  & 1 &  2 & 3	 & Order Average\\\hline 
Our method   & 35  & 49 & 60 & 7.1 \\
Random   & 6  & 8  & 11 & 23.1 \\
\end{tabular}
}
\label{table:exp2}
\end{center}
\end{table}

\subsection{Experiment 2: Estimation of the user's world model}
The query vector was applied to the obtained representation space to estimate the world model of others. 
In this verification, we assume that questions were asked in the situations of acquiring a key and opening a door, and the query ``In the state $S_{query}$, we should \{take the key/open the door\}'' was considered.
Figure \ref{fig:exp1-2} displays the results of estimating the others' world model given the reference world model and query. 
The environment that satisfies the query has the highest evaluation value, and the environment in which the key is located on the opposite side of the grid environment has the lowest evaluation value. 
This result suggests that the obtained query vector is appropriate for estimating the world model.

The optimal environment is defined as the agent's world model with minimal modifications for satisfying the query. 
For example, given the query ``In state $S_{query}$, the agent should take the key,'' the optimal environment is the environment in which only the position of the key is changed to satisfy the query, whereas the position of the door is left unchanged. 
For each randomly selected agent world model/query pair, the evaluation values for each environment obtained using Eq. (\ref{eval}) were sorted in descending order to obtain the order of appearance of the optimal environment (Table \ref{table:exp2}). 
The order of appearance of the environments that satisfy the query when sorted randomly is displayed for comparison.
The experimental results confirmed that the proposed method ranks the optimal environments higher.

Although the direct manipulation of the positions of keys and doors can change the order of the appearance of the optimal environment to one, direct manipulation is not always possible in cases in which directly manipulatable information is not given as a query. 
The proposed method estimates the optimal environment at an early stage, although the state transition relations to be changed are not explicitly given. 
This property of the proposed method is crucial.

\begin{figure}[t]
\centering
    \includegraphics[width=0.8\linewidth]{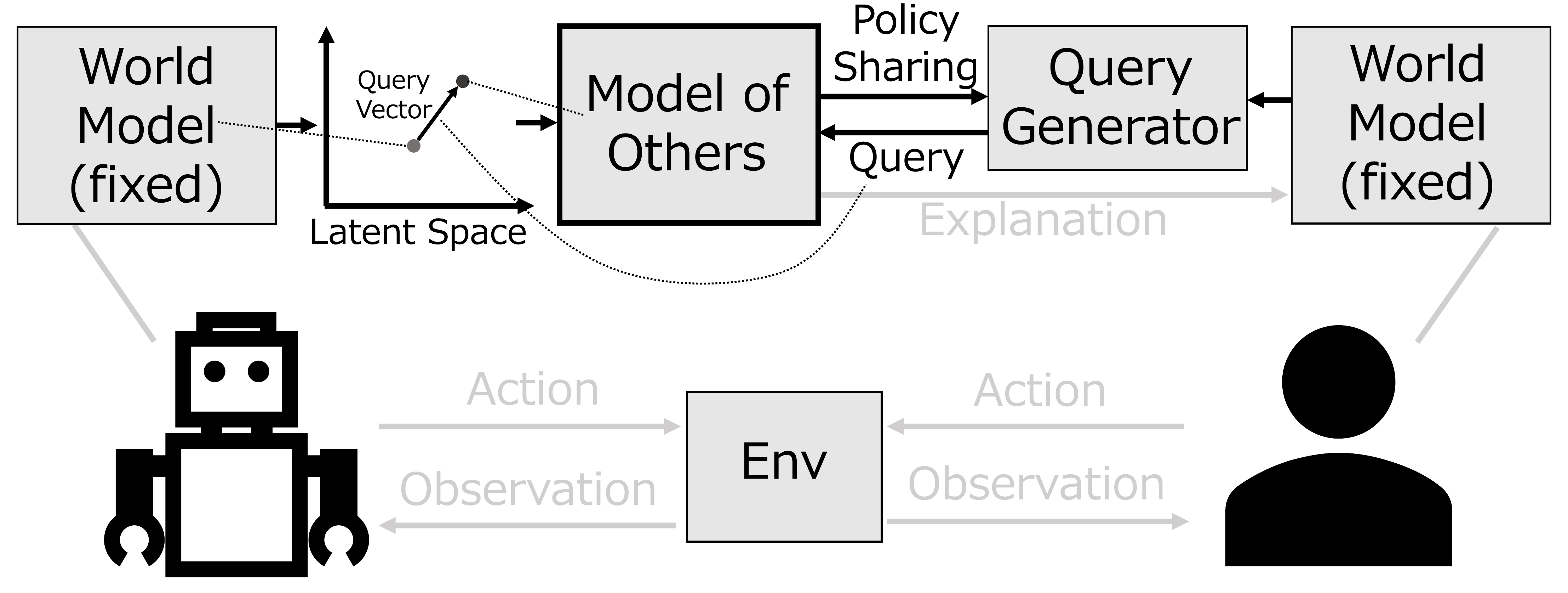}
    \caption{Schematic of experiment 3. The agent transmits information on its policy to the user and updates the user's world model based on queries.} 
    \label{fig:exp3}
\end{figure}

\subsection{Experiment 3: Validation with multiple queries}
An explanatory agent A and an explained agent B are prepared, and the number of queries required for A to accurately estimate B's world model is evaluated.
This experiment assumes a situation in which the user is asked to confirm the correctness of the estimation results of the other's world model through the presentation of an action sequence, which improves the estimation accuracy (Fig. \ref{fig:exp3}). 
The outline of the experiment is as follows:

\begin{itemize}
    \item[(1)]
    A and B share an initial state (absolute position and orientation of the agent, and the state of the key and door) and an environment-policy pair that specifies the strategy to be used in specific environments.
    They have arbitrary world models with different key and door positions.
    \item[(2)] 
    Here, A sets its own world model as the initial value of the other's world model and presents the optimal sequence of actions in that model in turn (policy sharing) \footnote{Policy sharing in this experiment (Fig. \ref{fig:exp3}) is performed to confirm the estimation results of the other's world model and differs from the presentation of information about one's own policy in Fig. \ref{fig:intro1} and Fig. \ref{fig:intro2}.}.
    \item[(3)] 
    B adds the query ``In state $s_{query}$, action $a_{query}$ should be chosen'' when the given action differs from the optimal action in its measure.
    The existing query is not deleted.
    \item[(4)] 
    A updates the other's world model based on the query and presents the action sequences in the updated other's world model again in sequence with $S_{query}$ as the initial state.
    \item[(5)] 
    Repeat (3) and (4) to evaluate the number of environment updates required for A to obtain B's world model as the other's world model. 
\end{itemize}

The environment selected once is not selected, and the second or subsequent candidate is adopted.
If the same environment has not been obtained after all the action sequences are presented, the environment is continuously updated without increasing the number of queries.
In this verification, the proposed method was compared with the “AND” search of queries as a method of updating the world model of others. 
In practice, the following three methods are compared.

 \begin{description}
  \item [Proposed method: ]\mbox{}Select a plausible environment using Eq. (\ref{others}). 
  \item [AND search 1: ]\mbox{}Randomly selects an environment from among the environments that satisfy the query. 
  \item [AND search 2: ]\mbox{}The environment is randomly selected with the constraint that ``the optimal behavior for the policy B is selected in all states from the initial state until reaching state $s_{query}$''. Thus, it adds constraints and increases the information provided compared with the two update methods described.
 \end{description}

 \begin{table}[t]
\begin{center}
\caption{Number of updates required to estimate the user's world model.}
\scalebox{1.0}{
\begin{tabular}{c|c | c }
   Method  & Number of updates &  Standard deviation\\\hline 
Our method   & {\bf 5.53}  & 4.83  \\
AND search 1   & 20.72  & 17.43   \\
AND search 2   & 8.69  & 4.71  \\
\end{tabular}
}
\label{table:exp3}
\end{center}
\end{table}
 
Experimental results revealed that the proposed method can estimate others' world models with the fewest number of updates (Table \ref{table:exp3}). 
The results of the corresponding two-tailed $t$-test revealed that $t(100)=8.07$ and $p<.01$ for the proposed method and AND search 1, and $t(100)=4.59$ and $p<.01$ for the proposed method and AND search 2. Thus, both significant differences were confirmed. 

The most common information given directly is ``AND search2''. 
By contrast, the proposed method can reduce the number of updates by vectorizing queries and utilizing prior knowledge embedded in the representation space as additional information.

\subsection{Experiment 4: Comparison with the use of probabilistic evaluation}
We compare the proposed method with the case where the probability value of each environment satisfying the query is used as the evaluation function instead of CAV. 
The proposed method uses Eq. (\ref{eval}) as the evaluation function, while the comparison method uses Eq. (\ref{eval_compare}).

\begin{equation}
\begin{split}
\label{eval_compare}
S(\bv_{i},\bv_{obs}, V_q)= {\sum_{j} P(a_{query}|\bv_{i},s_{query})} -\lambda \cdot distance(\bv_{i}, \bv_{obs}).
\end{split}
\end{equation}

The procedure for this experiment is the same as in experiment 3; however, the initial world model is not completely random, and the coordinates of either the key or the door are assumed to be identical. 
This condition replicates the assumption that the agent's world model and the user's world model are similar\footnote{Without this assumption, the world model with minimal modification to satisfy the query (the optimal environment) is not necessarily the user's world model. 
However, theoretically, when the evaluation value is calculated using Eq. (\ref{eval}), other environments that satisfy the query will have a lower evaluation value compared to the optimal environment. 
Therefore, if the assumption that the world models of the agent and user are similar cannot be made, it is desirable to use Eq. (\ref{eval_compare}). 
However, in a real environment, the world models of the agent and user are not completely independent, and similarity can be assumed.}.

\begin{table}[t]
\begin{center}
\caption{Comparison with the use of probabilistic evaluation.}
\scalebox{1.0}{
\begin{tabular}{c|c | c }
   Method  & Nuber of updates &  Standard deviation \\\hline 
Our method ($\lambda = 0.05$)   & {\bf 3.93}  & 5.38  \\
Our method ($\lambda = 0$)   & 5.75  & 7.11   \\
Probabilistic value ($\lambda = 0.05$)  & 7.13  & 7.79  \\
\end{tabular}
}
\label{table:exp4}
\end{center}
\end{table}

Experimental results showed that the proposed method, which takes into account the distance in the representation space ($\lambda = 0.05$), was able to estimate the user's world model with the fewest number of updates (Table\ref{table:exp4}). 
The results of the corresponding two-tailed $t$ test showed that the proposed method with $\lambda = 0.05$ compared to $\lambda = 0$ showed $t(99)=4.59$ and $p<.005$, while the proposed method with $\lambda = 0.05$ compared to the method using probability values showed $t(99)=4.44$ and $p<.005 $, both of which are significantly different from each other\footnote{Because the $t$ test was applied twice in this verification, a significant difference was found at the significance level $\alpha = 0.01$ based on the Bonferroni method.}.

The proposed method, which takes into account the distance in the representation space, was able to estimate the environment with a significantly smaller number of updates than the method using the same coordinates for either the key or the door. 
Compared to the method using probability value as the evaluation function, the proposed method was able to absorb small errors in probability values, resulting in a significantly smaller number of updates. 
When the distance in the representation space corresponds to the similarity of the state transition structure between environments, as in the present verification, it is effective to use CAV to obtain environments that are perpendicular to the virtual separation boundary as the user's world models.

\subsection{Experiment 5: Number of samples and accuracy of CAV}
We evaluate the number of queries required to correctly estimate the world model when the number of samples (number of environments) used to compute the CAV is reduced.
The evaluation procedure is the same as in Experiment 4. In this experiment, the maximum number of samples is 300 because 300 environments are embedded in the representation space. We also set $\lambda = 0.05$.

\begin{table}[t]
\begin{center}
\caption{Relationship between the number of CAV samples and the order of appearance of the optimal environment. Each value represents the cumulative frequency, and the number of trials is 100 for each.}
\begin{tabular}{c|c c  c}
   \diagbox{Nuber of samples}{Order of appearance}  & 1 &  2 & 3\\\hline 
300  & 40  & 62 & 69  \\
250  & 44  & 61 & 64\\
200  & 42  & 51 & 62  \\
150  & 40  & 51 & 55 \\
100  & 35  & 46 & 54 \\
50   & 9  & 18 & 26  \\
\end{tabular}
\label{table:exp5}
\end{center}
\end{table}

The experimental results show that the accuracy deteriorates much more slowly up to 100 samples than when the CAV is generated with 300 samples (Table \ref{table:exp5}).
This suggests that a specific level of estimation accuracy can be maintained even when the number of samples is reduced. 
The fact that the user's world model is estimated taking into account the distance in the representation space may also contribute to maintaining accuracy.
On the other hand, the accuracy dropped drastically when the number of samples was 50. 
This may be because of an increase in the number of trials in which the number of positive data (the number of data satisfying the query) in the sample is very small.

\subsection{Experiment 6: Use of the user vector}
\begin{figure}[t]
\centering
    \includegraphics[width=0.7\linewidth]{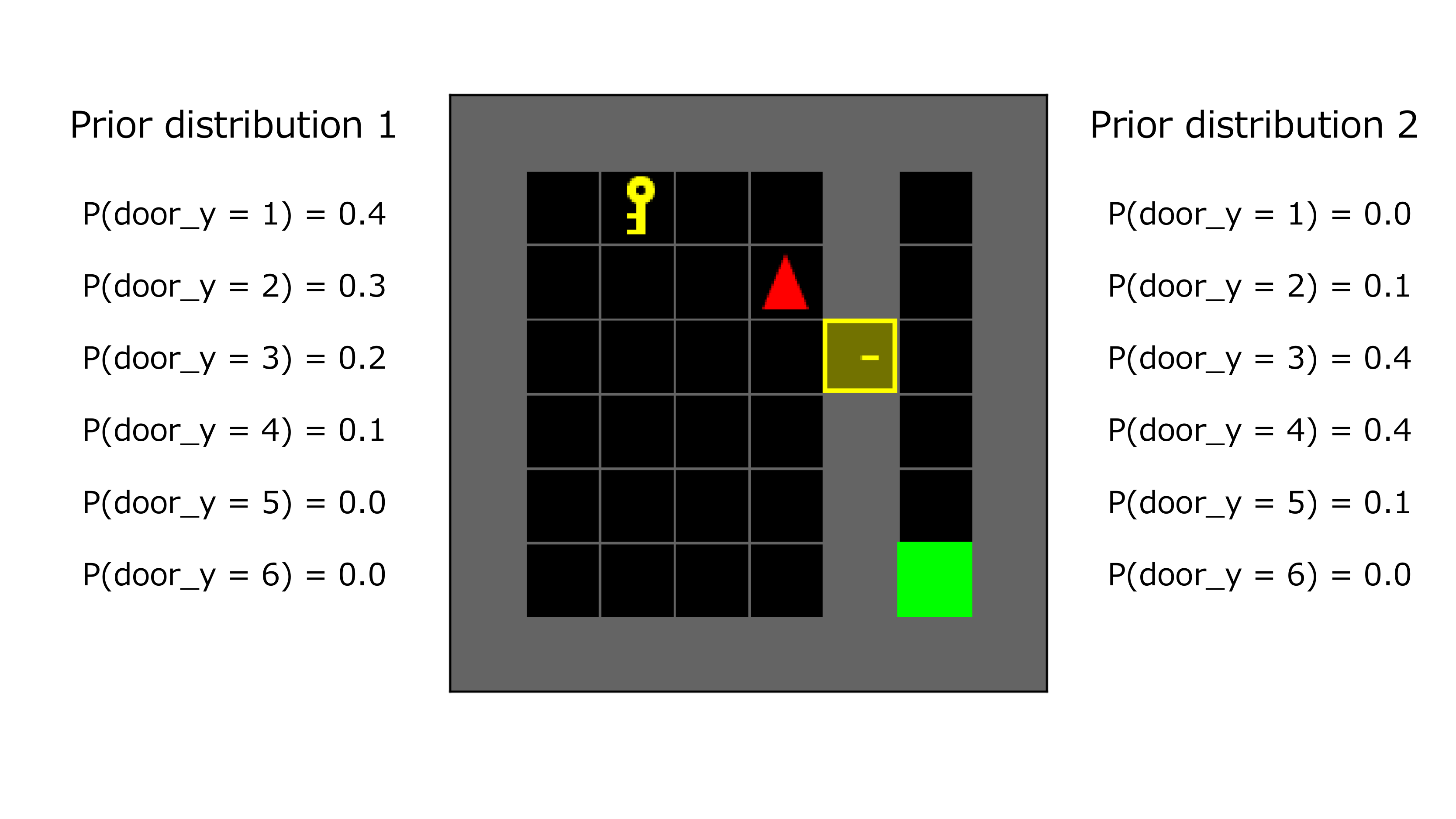}
    \caption{The prior distribution used in experiment 6. The prior distribution for the $y$-coordinate of the door (in the vertical direction) is defined.} 
    \label{fig:prior} 
\end{figure}

We test whether the estimation accuracy of the other-world model can be improved by using the user vector defined by Eq. (\ref{user_vec}), as opposed to using only CAV. 
In this verification, two types of prior distributions are defined for the y-coordinate of the door (Fig. \ref{fig:prior}), and the user vector is calculated for each of them. 
The user vector is treated as one of the query vectors in the evaluation value calculation for each environment. 
Note that $\lambda = 0$ is assumed in this verification because there is no assumption that the world models held by agents A and B are similar\footnote{If this experiment is conducted under the assumption that the world models of agents A and B are similar, it is necessary to set the number of objects that determine the state transition structure of the environment to three or more and that they are placed at the same coordinates as the current environment. Under these conditions, it is desirable to set $\lambda = 0.05$.}.

\begin{figure}[t]
\centering
    \includegraphics[width=0.7\linewidth]{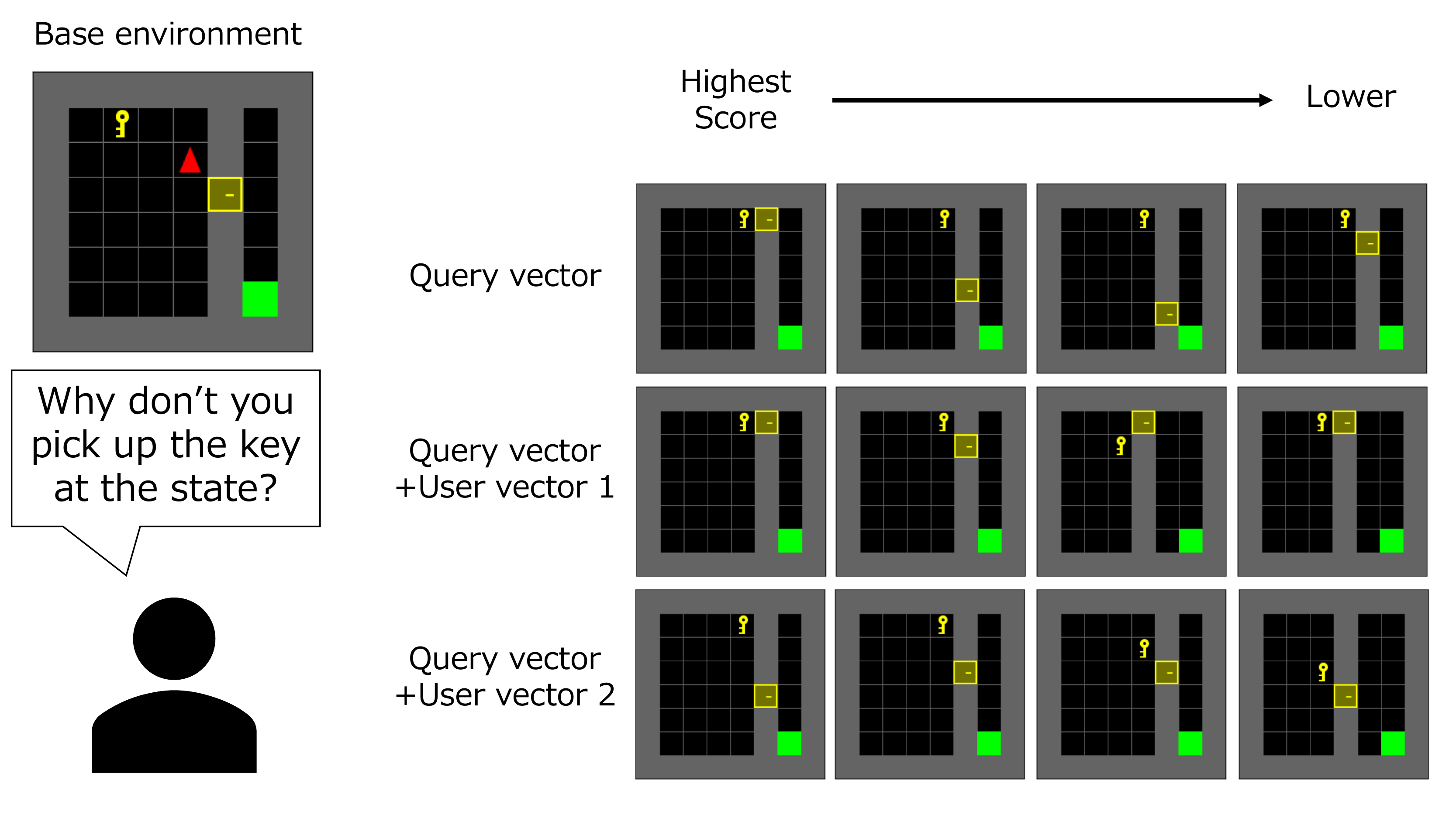}
    \caption{User vector and environments.} 
    \label{fig:exp6} 
\end{figure}

The estimation results of the other-world model for the same reference environment and query are shown in Fig. \ref{fig:exp6}.
User vectors 1 and 2 correspond to prior distributions 1 and 2, respectively.
It can be seen that while the results of the inference of the door location are unstable when only the query vector is applied, the results of the inference using the other vectors show that the door coordinates are concentrated in locations that have high probability values in the prior distribution.

\begin{table}[t]
\begin{center}
\caption{Change in estimation accuracy when using user vectors. The number of trials is 100 for each.}
\scalebox{1.0}{
\begin{tabular}{c|c c | c c}
    {} & \multicolumn{2}{c|}{Distribution 1} & \multicolumn{2}{c}{Distribution 2}\\
     Method & Number of updates & Standard deviation & Number of updates &  Standard deviation\\\hline 
Query + User  & {\bf 5.31}  & 4.59 & {\bf 4.69} & 3.34  \\
Query  & 6.39 & 7.13 & 5.11 & 3.46\\
\end{tabular}
}
\label{table:exp6}
\end{center}
\end{table}

The same validation as in experiment 3 was conducted by applying the prior distribution shown in Fig. \ref{fig:prior}, and the number of queries required for both distributions 1 and 2 was lower when using the user vector (Tab. \ref{table:exp6}). 
The results of the corresponding two-tailed $t$ test showed that $t(99)=2.34$ and $p<.05$ for distribution 1 and $t(99)=1.45$ and $p = 0.15$ for distribution 2. 
These results suggest that the number of queries required for estimation can be reduced by using the user vector, although the size of the effect depends on the shape of the prior distribution.

\subsection{Experiment 7: Explanation by language}
\label{section:mapping_exp}
We test whether the language vectors learned in advance can be used to correctly output language that describes the relationship between the agent's world model and user's world model. 
The language vectors used in this study are eight different explanations, such as ``In the world model assumed by the user, \{key, door\} is located at \{upper, lower, right, left\} than in the world model maintained by the agent.'' 
In the experiment, language explanations were first given to $n$ pairs of world models with different coordinates of keys or doors, and the language vectors were obtained using Eq. (\ref{word_vec}). 
We then generated linguistic explanations for randomly selected pairs of world models using the same conditions as those used to generate the language vectors and evaluated the percentage of the explanations that correctly explained the relationships between the world models (i.e., the percentage of correct responses).

The percentage of correct responses after 100 trials is shown in Table \ref{table:word_accuracy}. 
Note that if more than one linguistic explanation correctly represented the relationship between the world models, both were considered as correct answers. 
For example, if the key is located on the upper right, ``the key is on the right'' and ``the key is on the top'' are treated as correct answers. 
The ``1st'' in the table indicates the percentage of correct explanations generated by Eq. (\ref{relation}). 
The ``1st and 2nd'' represents the percentage of correctness of the two languages that were the first and second most similar. 
Note that ``1st and 2nd'' was evaluated only in cases where more than one language explanation was considered to be a correct answer. 

Although the number of world model pairs considered in this experiment is approximately 9000, the experimental results show that language explanation generation is possible with high accuracy even when the number of data used for language vector acquisition is $n=1000$. 
The accuracy was also maintained even when the number of data was extremely reduced to $n=100$ and $n=50$, suggesting that the learning in the representation space is effective. 
In this experiment, only eight language vectors were set, but it is expected that more accurate explanation generation will become possible by setting more detailed language vectors. 
However, it should be noted that in these cases, sufficient teacher data is required.

\begin{table}[t]
\begin{center}
\caption{Accuracy of language description generation.}
\scalebox{1.0}{
\begin{tabular}{c|c | c }
$n$ & 1st & 1st and 2nd \\\hline
5000  & 0.89  & 0.67 \\
3000  & 0.89 & 0.73 \\
1000  & 0.88 & 0.68 \\
500  & 0.84 & 0.63 \\
300  & 0.80 & 0.57 \\
100  & 0.69 & 0.54 \\
50  & 0.60 & 0.37 \\
\end{tabular}
}
\label{table:word_accuracy}
\end{center}
\end{table}

\section{Conclusion}
In this study, a novel method was proposed for estimating the user's world model from the robot's world model and the query given by the user to obtain XAR.
The proposed method can estimate others' world models more efficiently than using the “AND” search of queries. 
In the future, user vectors should be introduced, and the methods for generating explanations using differences in world models should be devised.

\section*{Acknowledgments}
This study was supported by the New Energy and Industrial Technology Development Organization (NEDO). 


\end{document}